\documentclass{article}

\usepackage{arxiv}
\usepackage{graphics}
\usepackage{graphicx}
\usepackage[utf8]{inputenc} 
\usepackage[T1]{fontenc}    
\usepackage{hyperref}       
\usepackage{url}            
\usepackage{booktabs}       
\usepackage{amsfonts}       
\usepackage{nicefrac}       
\usepackage{microtype}      
\usepackage{lipsum}
\def\etal{\textit{et al.}}
\usepackage{siunitx}
\usepackage{cite}
\usepackage{float}

\title{Unmanned Aerial Vehicles for Wildland Fires
}

\author{
  Moulay A.~Akhloufi\\
  Perception, Robotics, and Intelligent Machines Research Group (PRIME)\\ 
  Department of Computer Science\\
  Universit\'e de Moncton\\
  Moncton, Canada, E1A 3E9 \\
  \texttt{moulay.akhloufi@umoncton.ca} \\
  \AND
  Andy Couturier\\
  Perception, Robotics, and Intelligent Machines Research Group (PRIME)\\ 
  Department of Computer Science\\
  Universit\'e de Moncton\\
  Moncton, Canada, E1A 3E9 \\
  \texttt{eac6996@umoncton.ca} \\
    \And
  Nicol\'as A.~Castro\\
  Department of Electronic Engineering\\
  Universidad T\'ecnica Federico Santa Mar\'ia\\
  Valpara\'iso, Chile\\
  \texttt{nicolas.castro.12@sansano.usm.cl} \\
}

\begin{document}
\maketitle

\begin{abstract}
Wildfires represent an important natural risk causing economic losses, human death and important environmental damage. In recent years, we witness an increase in fire intensity and frequency.  Research has been conducted towards the development of dedicated solutions for wildland and forest fire assistance and fighting. Systems were proposed for the remote detection and tracking of fires. These systems have shown improvements in the area of efficient data collection and fire characterization within small scale environments. However, wildfires cover large areas making some of the proposed ground-based systems unsuitable for optimal coverage. To tackle this limitation, Unmanned Aerial Systems (UAS) were proposed. UAS have proven to be useful due to their maneuverability, allowing for the implementation of remote sensing, allocation strategies and task planning. They can provide a low-cost alternative for the prevention, detection and real-time support of firefighting. In this paper we review previous work related to the use of UAS in wildfires. Onboard sensor instruments, fire perception algorithms and coordination strategies are considered. In addition, we present some of the recent frameworks proposing the use of both aerial vehicles and Unmanned Ground Vehicles (UV) for a more efficient wildland firefighting strategy at a larger scale.
\end{abstract}

\keywords{Unmanned Aerial Systems \and UAS \and Autonomous Systems \and Wildfire \and Forest fires \and Fire Detection}

\vspace{4mm}

\textbf{\textit{Cite as:}}
Akhloufi, M.A.; Couturier, A.; Castro, N.A. "Unmanned Aerial Vehicles for Wildland Fires: Sensing, Perception, Cooperation and Assistance". Drones. 2021; 5(1):15; pp 1-25. DOI:10.3390/drones5010015

A recent published version of this paper is available at: \textbf{\underline{\url{https://doi.org/10.3390/drones5010015}}}
\vspace{1mm}

\section{Introduction}
\label{sec:intro}

Wildland and forest fires are an important threat in rural and protected areas.
Their control and mitigation are difficult as they can quickly spread to their surroundings, potentially burning large areas of land and getting close to urban areas and cities. 
The occurrence of wildfires results into substantial costs to economy, ecosystems and climate \cite{Jolly15}.
In western U.S. alone, wildfires increased by 400\% in the last decades \cite{Wildfires2019, Westerling2006}.
Kelly \etal \cite{Kelly2013} reported an increase in intensity and frequency of wildfires compared to the past 10,000 years.
In 2018, 8.8 million acres were burned by more than 58,083,000 wildfires \cite{IIS2019}. 
In Northern California, for example, the Camp Fire caused 85 deaths. This fire was the most destructive in California history burning 153,336 acres and destroying 18,733 structures. Losses were estimated to \textdollar16.5 billion \cite{Wildfires2019}. 
Experts estimate that wildfires will increase in the coming years mainly because of climate change \cite{CSSR2017}.

Two key elements are relevant in the effort to reduce the impact of a wildfire. 
First, the time span between the start of a fire and the arrival of firefighters which needs to be reduced to a minimum. 
Thus decreasing the chances of fire to spread to unmanageable levels.
Second, the severity evaluation of the event and the monitoring of the emergency response which is necessary to elaborate better fighting strategies.
For these reasons, it is important to have reliable and efficient systems for early stage fire detection and monitoring.

Research efforts have been devoted to the development of systems that can cope with this type of scenario.
Remote sensing is essential to this field because it prevents humans to be unnecessarily exposed to dangerous activities.
Satellite images have been used for the detection of active fires \cite{Chiaraviglio16,Fukuhara17} and the reporting of fire risks \cite{Chien11}.
Wireless sensor networks (WSNs) have also been proposed for wildfires detection \cite{Yoon12}, monitoring \cite{Tan11} and risk assessment \cite{Lin17}.
Both types of systems have practical limitations.
In the first approach, depending on the image resolution, the data is averaged over pixels making it difficult to detect small fires \cite{NASA2019}.
Furthermore, satellites have limited ground coverage and it can take a significant amount of time between two passes over the same point, making them unsuitable for continuous monitoring. 
In the case of WSNs, the maintenance difficulties, the lack of power independence and the fact that they are not scalable (due to their static nature) severely reduce their coverage capacities \cite{WSN_Limit2013}.
Moreover, in case of fire the sensors are destroyed leading to additional replacement costs.
Unmanned Aerial Systems (UAS) are more convenient for this task due to their maneuverability, autonomy and easy deployment.

UAS have known impressive progress and are widely used today in various fields. They become smaller, more affordable and also gained higher processing capabilities turning them into reliable tools for remote sensing missions in hostile environments.
Research has shown their benefit in technical and strategical assistance for surveillance, monitoring, and tasks related to post-fire damage evaluation \cite{Ollero06, Ambrosia11, Hinkley11, Yuan15a, Skorput16}.
Additionally, UAS have exhibited a positive economic balance in favor of their use in wildfire emergencies \cite{Restas06, Laszlo18} making them both practical and economically reliable.
UAS can fly or hover over specific zones to retrieve relevant data with cameras or other airborne sensors.
This has motivated authors to propose frameworks and techniques using UAS with the goal of delivering optimal fire detection, coverage and firefighting.
This paper presents a survey of the main literature related to the use of UAS in the context of wildland and forest fires.
We review the fire assistance architectures proposed, the sensing modalities used, the fire perception approaches ad datasets and the UAS coordination strategies.

\section{Fire assistance systems architecture}
\label{sec:overview}

Remote sensing with aerial systems present multiple advantages in the context of emergency assistance.
Their high maneuverability allows them to dynamically survey a region, follow a defined path or navigate autonomously.
The wide range of sensors that can be loaded onboard allows the capture of important data which can be used to monitor a situation of interest and plan the emergency response.
The ability to remotely control an UAS helps reduce the risk for humans and remove them from life-threatening tasks.
The automation of maneuvers, planning and other mission-related tasks through a computer interface improves distant surveillance and monitoring.
Advances in these aspects have a direct impact on the firefighting resource management.

General purpose fire assistance systems can be divided into three main components: sensing modalities and instruments, fire perception algorithms, and coordination strategies between multiple UAS or with the Ground Control Station (GCS).
These components are designed to perform one or more tasks related to fire emergencies.
Within the reviewed works, the most developed aspects are fire detection and monitoring.
Fire prognosis and firefighting, although they are present in some work, are less prominent.
Fire detection and monitoring is centered around recognition techniques, a field of research that has seen significant advances in the last decades.
Fire prognosis and fighting has practical limitations that hinder research in these areas.
Prognosis requires complex mathematical models that must be fed with data that can be difficult to acquire in real-time and in unknown environments.
Fire fighting, on the other hand, requires expensive combat equipment especially for large wildland fires.
Furthermore, close proximity with fires can pose a significant risk for the vehicle integrity and lead to its loss.
Research has been done to design a UAS capable of fighting fires \cite{Twidwell16, Qin16} and more recently some drones manufacturers have step in to tackle this problem \cite{DJIFire2019}, but more work remains to be done for these vehicles to be affordable and technically viable.

UAS have different sizes, maneuverability, and endurance capacities.
There is a wide selection of aerial systems ranging from large UAS with long endurance and high-processing capabilities to small UAS with short flight times and limited processing capabilities \cite{UAVClass2012, UAVClass2019}.
Large vehicles are expensive but have higher payload and can carry more sensors and other instruments. In the other hand, smaller vehicles are more affordable but with limited payload.
The instruments onboard the vehicles vary between the reviewed systems, but some are essential to navigation and localization and therefore found in almost all UAS.
Global Navigation Satellite Systems (GNSS) and Inertial Navigation Systems (INS) falls into this last category.
Furthermore, almost all of the vehicles have at least one kind of imagery sensor used for different purposes including fire perception.
Temperature sensors are also present in some of the proposed fire assistance systems, but as shown in section \ref{sec:instruments}, they are less popular.

Sensor measurements are the input of the fire perception algorithms that process the data to detect the presence of the fire. 
The processing can be either onboard the UAS or in a computer located at a GCS. 
The fire perception can also be performed by a human operator inspecting the data from a GCS.  
Most of the developments are devoted to the automation of fire perception and the optimization of the processing while preserving the accuracy. 
Computer vision and machine learning techniques are widely used for this purpose.

The last component is the coordination strategy, it provides the framework for the deployment of the flight mission.
Surveillance missions are usually planned beforehand and aim to search wide areas, prioritizing areas with higher fire risks.
The coordination strategy becomes essential during the monitoring of a fire propagation, as it is necessary to adapt the flight plan to the fire spread.
This is even more relevant if there are multiple UAS collaborating to the mission during a fire emergency.
Multiple coordination strategies were proposed. 
For example, one UAS could hover near a fire spot and alert the rest of the fleet to proceed with fire confirmation \cite{Martinez-de-Dios07}.
More complex planning is also possible, by requiring a consensus on the task to be performed by each unit \cite{Ghamry17} or by flying in a specific formation around the fire perimeter \cite{Casbeer06}.
In both cases, a concrete description of the task and the autonomous decision scheme must be defined.
Section \ref{sec:coord} gives more details about the coordination strategies using single or multiple UAS.

Table \ref{tab:review} presents an overview of the main fire assistance systems.
The first column contains the reference of the work.
The second column presents the sensing modalities used to perform fire perception.
Some of the systems do not specify sensing instruments and the authors assume that the necessary instruments are available.
The third column describes the tasks that are addressed in the corresponding research.
The possible tasks are detection, monitoring, diagnosis, prognosis and fighting.
The fourth column gives the used coordination strategy,  specifying if it is a man-controlled vehicle, a single UAS, or a fleet of vehicles.
The last column describes the validation methodology used in each work.
It can be either by system overview, simulation, laboratory tests (near practical), or real flight tests (practical).


\begin{table}[h!]
\caption{Characteristics of fire assistance systems.}
\begin{center}       
\begin{tabular}{|p{3.2cm}|p{2.75cm}|p{2.75cm}|p{3.2cm}|p{2cm}|}

\hline
\rule[-1ex]{0pt}{3.5ex}  \textbf{Authors}  &  \textbf{Sensing mode} & \textbf{Tasks} & \textbf{Coordination} & \textbf{Validation}  \\
\hline
\rule[-1ex]{0pt}{3.5ex}Casbeer \etal \cite{Casbeer06}  & IR & Monitoring & Centralized team & Simulation  \\
\hline
\rule[-1ex]{0pt}{3.5ex}Bradley \etal \cite{Bradley11}  & IR & Detection and monitoring & Single UAS & Near practical  \\
\hline
\rule[-1ex]{0pt}{3.5ex}Hinkley \etal \cite{Hinkley11}  & IR & Monitoring & Single man-controlled UAS & Practical  \\
\hline
\rule[-1ex]{0pt}{3.5ex}Kumar \etal \cite{Kumar11}  & IR & Monitoring and fighting & Collaborative team & Simulation  \\    
\hline
\rule[-1ex]{0pt}{3.5ex}Pham \etal \cite{Pham17}  & IR & Monitoring & Collaborative team & Simulation  \\
\hline
\rule[-1ex]{0pt}{3.5ex}Yuan \etal \cite{Yuan17a}  & IR & Detection & Centralized team & Simulation  \\
\hline
\rule[-1ex]{0pt}{3.5ex}Belbachir \etal \cite{Belbachir15} & Temperature & Detection & Multiple UASs. No collaboration & Simulation   \\
\hline
\rule[-1ex]{0pt}{3.5ex}Lin \etal \cite{Lin15}  & Temperature & Monitoring & Collaborative team & Simulation  \\
\hline
\rule[-1ex]{0pt}{3.5ex}Yuan \etal \cite{Yuan15b,Yuan16a,Yuan16b,Yuan17b} & Visual & Detection & Single UAS & Near practical  \\
\hline
\rule[-1ex]{0pt}{3.5ex}Sun \etal \cite{Sun17}  & Visual & Detection and Monitoring & Single man-controlled UAS & Validation  \\
\hline
\rule[-1ex]{0pt}{3.5ex}Merino \etal \cite{Merino05,Merino06}, Martinez-de-Dios \etal \cite{Martinez-de-Dios07} & IR and Visual & Detection and Monitoring & Centralized team & Practical  \\
\hline
\rule[-1ex]{0pt}{3.5ex}Martinez-de-Dios \etal \cite{Martinez-de-Dios11} & IR and Visual & Monitoring and diagnosis &  Single UAS & Practical  \\
\hline
\rule[-1ex]{0pt}{3.5ex}Pastor \etal \cite{Pastor11} & IR and Visual & Detection and Monitoring & Single UAS &  System overview  \\
\hline
\rule[-1ex]{0pt}{3.5ex}Merino \etal \cite{Merino12,Merino15} & IR and/or Visual & Detection and Monitoring & Collaborative & Practical  \\
\hline
\rule[-1ex]{0pt}{3.5ex}Martins \etal \cite{Martins07}  & NIR and Visual & Detection & Single UAS & Simulation  \\
\hline
\rule[-1ex]{0pt}{3.5ex}Ambrosia \etal\cite{Ambrosia11} & Multispectral camera & Detection and diagnosis & Single man-controlled UAS & Practical  \\
\hline
\rule[-1ex]{0pt}{3.5ex}Sujit \etal \cite{Sujit07} & Not specified & Monitoring& Collaborative team & Simulation  \\
\hline
\rule[-1ex]{0pt}{3.5ex}Alexis \etal \cite{Alexis09}  & Not specified & Monitoring & Collaborative team & Simulation  \\
\hline
\rule[-1ex]{0pt}{3.5ex}Karma \etal \cite{Karma15}  & Not specified & Monitoring & Single man-controlled UAS & Practical  \\
\hline
\rule[-1ex]{0pt}{3.5ex}Ghamry \etal \cite{Ghamry16c,Ghamry16d}  & Not specified & Detection and monitoring & Collaborative team & Simulation  \\ 
\hline
\rule[-1ex]{0pt}{3.5ex}Ghamry \etal \cite{Ghamry17}  & Not specified & Fighting & Centralized team & Simulation  \\
    
\hline

\end{tabular}
\end{center}
\label{tab:review}
\end{table}

\section{Sensing Instruments}
\label{sec:instruments}

Sensors provide the necessary data for navigation and for firefighting monitoring and assistance. 
In outdoor scenarios, GNSS and INS provide real-time UAS localization.
They are also used to georeference the captured images thus allowing geographical mapping of fires.
Fires have specific signatures that can be composed of different elements such as heat, flickering, motion, brightness, smoke and bio-product \cite{Allison16}.
These elements can be measured using suitable sensing instruments.
Cameras are the sensing instruments that offer the most versatility.
Visual and infrared (IR) sensors onboard UAS can be used to capture a rich amount of information.
Table \ref{tab:spectrum}, gives the main camera spectral bands used in fire studies. 

\begin{table}
\caption{Visual and IR electromagnetic spectrum.}
\begin{center}       
\begin{tabular}{|p{7.5cm}|p{5cm}|} 

\hline
\rule[-1ex]{0pt}{3.5ex}  \textbf{Spectral band}  &  \textbf{Wavelength $(\mu m)$ }  \\
\hline
\rule[-1ex]{0pt}{3.5ex}  Visible   &  0.4 - 0.75  \\
\hline
\rule[-1ex]{0pt}{3.5ex}  Near Infrared (NIR) & 0.75 - 1.4  \\
\hline
\rule[-1ex]{0pt}{3.5ex}  Short Wave IR (SWIR)& 1.4 - 3  \\
\hline
\rule[-1ex]{0pt}{3.5ex}  Mid Wave IR (MWIR)& 3 - 8  \\
\hline
\rule[-1ex]{0pt}{3.5ex}  Long Wave IR (LWIR)& 8 - 15  \\
\hline

\end{tabular}
\end{center}
\label{tab:spectrum}
\end{table} 

Visual, IR and other sensors proposed in past work are presented in the following. 

\subsection{Infrared spectrum}

The infrared electromagnetic spectrum wavelengths ranges from $0.7\mu m $ to $1000\mu m$ (Table \ref{tab:spectrum}).
Specialized sensors are available to capture images in different sub-bands of the IR spectrum.

All matter emits an electromagnetic radiation in some wavelength band which is proportional to the fourth power of the object absolute temperature.
At room temperature, the radiation peak is located within the thermal infrared band.
For a wildfire, temperatures can top the $1000$\si{\degree}\text{C} ( $1800$\si{\degree}\text{F}), leading to a peak radiation in the mid-wave infrared (MWIR) sub-band \cite{Johnston14,Allison16}.
Therefore, a sensor operating in the MWIR is best suited for fire perception.
However, until recently \cite{MWIR_FLIR2018} the form factor of MWIR sensors and their cost limited their use in low-cost small and medium UAS. 

\begin{figure} 
    \centering
    \includegraphics[width=0.8\columnwidth]{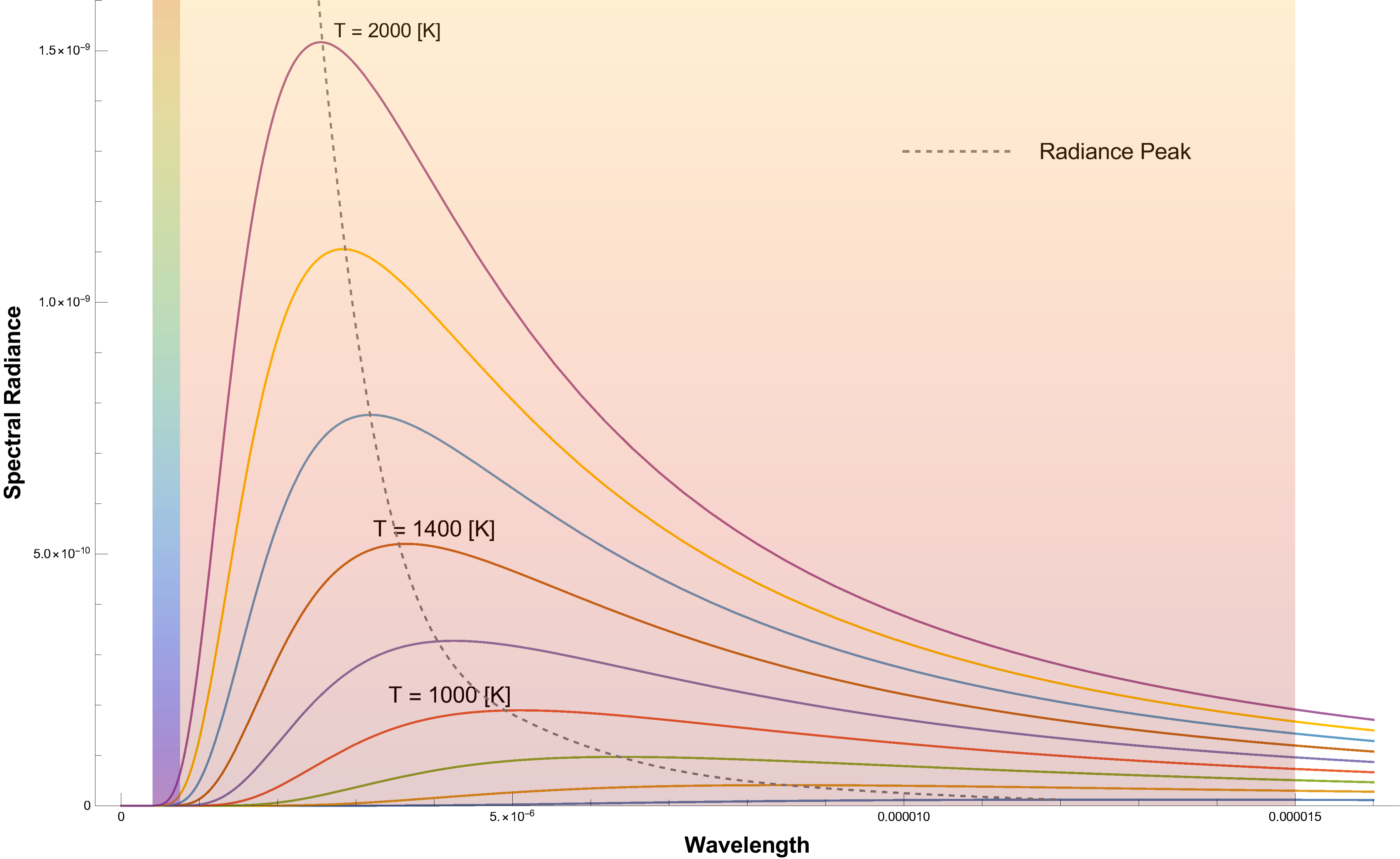}
    \caption{Black body radiance at different wavelengths. Each curve represents a specific temperature in Kelvin, starting at 200[K] at the bottom and up to 2000[K] at the top. Visible light and IR wavelengths are highlighted at the left and right, respectively.}
    \label{fig:spectrum}
\end{figure}

To overcome these restrictions in smaller aerial vehicles, fire detection systems are using mostly NIR, SWIR or LWIR sensors.
The use of these sub-bands is possible due to the fact that the higher temperature shifts the distribution of the object radiation to shorter wavelengths. This property can be seen in figure \ref{fig:spectrum} illustrating how the radiance changes as a function of the temperature.
It is not absolutely necessary to detect fire at the peak as the effect is also clearly visible in other close spectral bands.
However, a disadvantage of using NIR and SWIR is that objects under sunlight are often reflecting radiation in these sub-bands creating false positives.
Still hot fire spots remain detectable but the fire contrast is reduced during day time flights \cite{Allison16}.

Regardless of their characteristics, IR sensors are the most commonly used sensors in fire assistance systems due to their ability to detect heat. In
\cite{Bradley11}, \cite{Casbeer06}, \cite{Hinkley11}, \cite{Kumar11} and \cite{Yuan17a} the authors proposed methods based solely on IR spectrum (Table \ref{tab:review}).

\subsection{Visible spectrum}

Visible spectrum cameras are widely available and commonly used in various applications.
They come in a wide variety of resolutions, form-factors and cost.
Their versatility offers a valuable alternative in wildfires research from both technical and commercial perspectives.
Moreover, the ever continuing reduction in visible cameras size and weight makes them perfect candidates for UAS.

Data provided by these sensors are images in grayscale or RGB format.
This allows the development of computer vision techniques using color, shape, temporal changes and motion in images or a sequence of images.
Some of the vision-based techniques are presented in section \ref{sec:fireseg}.
Although, they are versatile and widely available, visible light sensors must be carefully selected for night time operations as some sensors perform poorly in low light conditions.
Despite some of their limitations, they equip almost all UAS and make them good candidates for wildland fires study. 
Authors in \cite{Yuan15b}, \cite{Yuan16a}, \cite{Yuan16b}, \cite{Yuan17b},  and \cite{Sun17} proposed systems that rely only on the use of visible spectrum cameras (Table \ref{tab:review}).

\subsection{Multispectral cameras}

Using each spectral band alone comes with its limitations. To tackle these limitations authors propose the use of multiple cameras and combine multiple spectrum. 
This allows the use of data fusion techniques to increase the accuracy of fire detection in complex situations and under different lighting conditions.
Esposito \etal \cite{Esposito07} developed a multispectral camera operating in the LWIR, NIR and visible spectrum mounted on a UAS.
In a NASA Dryden's project, Ikhana UAS \cite{Ambrosia11, NASAIkhana2007}, a Predator B unmanned aircraft system adapted for civilian missions, was built to carry a multispectral sensor that operates in 16 different bands from visible to LWIR spectrum.
Despite their interesting characteristics, in both cases, the weight of these combined sensors limited their implementation to large airborne platforms only.
To address this problem, other alternatives combine smaller sensors such as visible spectrum sensors and IR sensors.
In \cite{Martinez-de-Dios07}, the authors used this approach to capture and project the IR data onto visible images.
This generated a superposition of the data leading to pixels being represented with four intensity values red, green, blue and IR.
The authors reports improvements in fire detection with mixed segmentation techniques that make use of the four-channel values.

\subsection{Other sensors}

Various sensors other than cameras have been proposed to detect and confirm the presence of fires.
Some authors proposed the use of chemical sensors which can detect concentrations of hazardous compounds \cite{Karma15}.
Spectrometry measures can also be used to detect the characteristics of burned vegetation and confirm a fire \cite{Allison16}.
Again, in both cases, the size of the sensors limited their use.
Temperature sensors have also been used by Belbachir \etal \cite{Belbachir15} and Lin \etal \cite{Lin15} to generate heat maps and detect/locate fires.
\\

The remaining of the reviewed literature will be based on the used of cameras to detect, monitor and measure fires.

\section{Fire perception}
\label{sec:fireseg}

Research has shown the effectiveness of UAS as a remote sensing tool in firefighting scenarios \cite{Ambrosia11, Karma15, Hinkley11}.
They are very useful even in simple tasks such as observing the fire from a static position and streaming the video sequence to human operators.
This simple use case already allows firefighters to have an aerial view of the spreading fire and plan containment measures.
However, single man-controlled UAS, even if they are useful for small emergencies, do not scale up in large scenarios.
Automation of detection and monitoring of fires can help deliver an optimal coverage of the fire area with the help of multiple UAS.
Furthermore, the gathered data can be processed to analyze the fire, estimate its Rate of Spread (RoS) \cite{Ghamry16b}, volume \cite{Martinez-de-Dios11} or perform post fire damage evaluation \cite{Ambrosia11}.

To perform fire-related tasks autonomously, systems must address different tasks such as fire location estimation and path planning. They also need to dynamically adapt to the evolving emergency.
For that purpose, sensor data are processed to detect fire and extract fire-related measures.
For fire detection, authors usually extract fire-like pixels based on color cues or IR intensities.
Monitoring tasks usually require further analysis to estimate the fire perimeter or burned areas.
Computed measures can be used as input to fire models to estimate the fire propagation over time.
This section reviews some fire detection and segmentation techniques found in the literature.

\subsection{Fire segmentation}
Fire segmentation is the process of extracting pixels corresponding to fire in an image.
The criteria by which a pixel is selected vary from one method to another.
The selection criteria are also the main factor affecting the accuracy of the detection.
In general, fire segmentation uses the pixel values of a visual spectrum image (e.g. color space segmentation) or the intensities of an IR image.
Motion segmentation can also be used to extract the fire using its mouvement over a sequence of images.

\subsubsection{Color segmentation}

Images are built of pixel units that can have different encoding (e.g. grayscale, color). 
In color images, pixels are composed of three values in the red, green and blue channels (RGB).
Other color spaces are also possible such as YCbCr, HSI, CIELAB, YUV, etc. \cite{Fairchild2013}.
In IR images the pixels have a one channel value representing temperature (MWIR and LWIR) or reflectance (NIR, SWIR).

In the COMETS project \cite{Martinez-de-Dios07, Merino05, Merino06}, the authors use a lookup table with fire-like colors (RGB values) that were extracted from a learned fire color histogram.
The image pixels are compared to the table and the values that are not found are considered as non-fire.
A non-calibrated LWIR camera is used to capture qualitative images with radiation values relative to the overall temperature of the objects in the scene.
The heat peak observed in the resulting image depends on the current scenario.
A training process was carried out to learn the thresholds to be applied to the IR images for binarization.
Images with and without fires were considered as well as different lighting conditions and backgrounds.
This permitted the selection of the appropriate threshold to apply during deployment in known conditions.
Ambrosia \etal \cite{Ambrosia11} selected fixed thresholds for each IR spectral band.
They also varied the bands used for day and night missions.
During night time, the MWIR and LWIR bands were used and during the day the NIR band was added.
The results shows that fixed threshold adapt poorly to unexpected conditions but can be tuned to perform better in known environments.

In \cite{Yuan17a, Yuan15b,Yuan16a,Yuan16b,Yuan17b}, the authors use color space segmentation. The images are converted from RGB to the CIELAB color space prior to further processing.
Sun \etal \cite{Sun17} propose the use of YCbCr color space.
In both cases, a set of rules were developed based on empirical calculations performed on captured fire images.
For example, the authors in \cite{Sun17} considered pixels as fires if their values followed the following rules: $Y>Cb$, $ Cr>Cb$, $Y>Y_{mean}$, $Cb<Cb_{mean}$ and $Cr>Cr_{mean}$.
The mean sub-index indicates the channel mean value of the corresponding image.
Otsu thresholding technique \cite{otsu79} was used in \cite{Yuan17a} to segment IR images.

Color value rule-based segmentation approaches are computationally efficient, but lack robustness during detection.
Results show that objects with a color similar to fire are often mislabeled as fire and trigger false alarms.
A combination of rules in different color spaces and the addition of IR can increase the detection accuracy.
More complex algorithms that are time and space aware have also been shown to increase the accuracy of the fire detection \cite{lei_new_2017, wang_new_2017, chou_block-based_2017, shi_video-based_2016, abdullah_position_2016,steffens_texture_2016, choi_patch-based_2015, poobalan_fire_2015, zhang_fire_2015, toulouse2015benchmarking, verstockt_wavelet-based_2011}. But the majority of them have not been integrated with UAS. 

In recent years, deep learning algorithms have shown impressive results in different areas. In UAS, past work using deep convolutional neural networks (CNN) dealt mainly with fire detection \cite{FireDetectCVPR2019, UAVdeep2018, UAVdeep2017}. Deep fire segmentation techniques proposed recently has shown the potential of developing an efficient wildfire segmentation system \cite{akhloufi2018DeepFire}. The used dataset in this last work included some aerial forest fires images \cite{TOULOUSE2017dataset}. Deep segmentation of wildland fires is still lacking in UAS applications. 


\subsubsection{Motion segmentation}

Fire segmentation using static images help reduce the search space, but often objects with a similar color to fire can be detected and lead to false positives.
In \cite{Yuan17a, Yuan15b,Yuan16a,Yuan16b,Yuan17b} and \cite{Sun17}, the authors proposed the use of Lukas-Kanade optical flow algorithm \cite{Bruhn05} to consider fire movements.
With the detection of corresponding feature points in consecutive image frames, a relative motion vector can be computed.
The mean motion vector matches the UAS motion except for moving objects in the ground.
Fire flames are among those objects because of their random motion.
By detecting feature points within regions with both random movements and fire-like colors, the fire can be confirmed and the false alarm rate reduced.

\subsection{Fire detection and features extraction}

The data fed to a detection system are analyzed in order to find patterns that confirm the occurrence of an event.
Patterns are recognized by computing different features which can be strong or weak signatures for a specific application.
In the case of fire detection with UAS, the most popular features are color, brightness and motion.
Research focusing on fire detection considers the fusion of more features to obtain better results in the classification stage.
These features can be categorized by the level of abstraction at which they are extracted: pixel, spatial and temporal.

Color cues are widely used at the first step to extract fire-like pixels.
This reduces the search space for further processing with more computationally expensive detection algorithms.
For example, the RGB mean values of a Region of Interest (RoI) and the absolute color differences ($|R-B|$, $|R-G|$, $|B-G|$) can be thresholded \cite{asatryan_method_2015} or used to train a classification algorithm \cite{yuan_learning-based_2018}.
In the work of \cite{duong_efficient_2013}, the authors further added the intensity mean, the variance and the entropy values of the ROI to the feature vector.
Other features used in the literature include color histograms of ROI \cite{shi_video-based_2016} and color spatial dispersion measures \cite{wang_new_2017}.

After the detection of the ROI, other features can be extracted.
Some authors consider spatial characteristics to determine the fire perimeter complexity by relating the convex hull to the perimeter ratio and the bounding rectangle to perimeter ratio \cite{zhou_analysis_2015}.
The distance between the blob centroid position within to bounding box has also been considered in this work.

Texture is another spatial characteristic often used for fire detection.
The main texture descriptors proposed for this task are Local Binary Patterns (LBP) \cite{chen_fire_2014, chino_bowfire:_2015, chi_real-time_2017, favorskaya_verification_2015} and Speeded Up Robust Features (SURF) \cite{choi_patch-based_2015, shi_video-based_2016}.
These operators characterize local spatial changes in intensity or color in an image and return a feature vector that can be used as input for classification.
SURF \cite{bay2008_surf} is computationally expensive but allows for scale and rotation invariant matching.
LBP \cite{LBP96} needs less processing power and extract the mean relation between pixels in a small area using the 8 neighbors of a pixel.
Some authors \cite{avalhais_fire_2016, li_fire_2016} also use the Harris corner detector \cite{Harris88}, which is a computationally efficient feature point extractor.

The features reviewed above are extracted from single images. 
When a video sequence is available, the temporal variation in color, shape and position of some blobs can be extracted.
In \cite{ko_fire_2014}, the fire blob shape variation is computed by a skewness measure of the distance from the perimeter points to the blob’s centroid.
Foggia \etal \cite{foggia_real-time_2015} measured shape changes by computing the perimeter to area ratio variation over multiple frames.
The authors also detected the blob movements by matching them in contiguous frames and to compute the centroid displacement.
Fire tend to moves slowly upwards, thus blobs that do not comply with this rule can be discarded \cite{ko_fire_2014, lei_new_2017, buemi_efficient_2016}.
The centroid displacement can also be an input for further classification \cite{zhou_analysis_2015, lei_new_2017, cai_intelligent_2016}.
A similarity evaluation is employed in \cite{zhou_analysis_2015}. They measure the rate of change of overlapping area of blobs in contiguous frames.
This gives a practical representation of the speed at which the region of interest is moving and if it is growing or decreasing in size.
Fire flickering can also be identified by considering specific measures such as intensity variation \cite{zhao_hierarchical_2015}, the number of high-pass zero crossing in the wavelet transform \cite{stadler_comparison_2014} or the number of changes from fire to non-fire pixels inside a region \cite{barmpoutis_real_2013}. 
Wang \etal \cite{wang_spatial-temporal_2013} implemented a long-term movement gradient histogram, which accumulates the motion changes.
The histogram is fitted to a curve which is used to evaluate if the area corresponds to a fire or not.
Kim \etal \cite{kim_fire_2016} proposed a Brownian motion estimator that measures the correlation of two random vectors \cite{szekely2009}.
The vectors are composed of channel values, the first intensity derivative and the second intensity derivative.
Therefore, the Brownian motion estimator describes the dynamic dependence between a series of regions across multiple frames.
Temporal features consider a time window for the fire evaluation. Then, some empirical criteria are established to determine the optimal thresholds and duration of the events in order to trigger a fire alarm.

Among the features described so far, there are some features that are more oriented towards fire detection.
Features such as color, blob centroid displacement and flickering are some of the most popular.
Some novel approaches such as the Brownian correlation or the histogram of gradients have been less explored but are nevertheless interesting.
A comparison of these different features and an evaluation of which one has a greater impact on the fire detection accuracy and false positive rate would be very useful.
Unfortunately, to our knowledge, such a comprehensive comparison has not been yet published.
However, as most of these features are not computationally expensive, ensembling the features can improve the performance and reduce the false detection rate. Table \ref{tab:inputs} gives an overview of the features used depending on the input. 

\begin{table}[h!]
\caption{Image input and extracted features.}
\begin{center}       
\begin{tabular}{|p{2.8cm}|p{3.5cm}|p{3.5cm}|p{4cm}|} 

\hline
\rule[-1ex]{0pt}{3.5ex}  \textbf{Input}  &  \textbf{Statistical measures} & \textbf{Spatial features}  &  \textbf{Temporal features}   \\
\hline
\rule[-1ex]{0pt}{3.5ex}Color, IR and radiance images & Mean value, mean difference, color histogram, variance and entropy. & LBP, SURF, shape, convex hull to the perimeter rate, bounding box to the perimeter rate. & Shape and intensity variations, centroid displacement, ROI overlapping, fire to non-fire transitions, movement gradient histogram, and Brownian correlation.  \\

\hline
\rule[-1ex]{0pt}{3.5ex}Wavelet transform & Mean energy content. & Mean blob energy content. & Diagonal filter difference. High-pass filter zero-crossing of wavelet transform on area variation. \\

\hline
\end{tabular}
\end{center}
\label{tab:inputs}
\end{table}

\subsection{Considerations in UAS applications}

Additional features can improve the fire detection. 
Features that are obtained by temporal analysis evaluate the difference between contiguous frames.
In simple scenarios, where the camera is static and the background is not complex, frame subtractions can help detect moving pixels.
In the presence of complex and dynamic backgrounds, Gaussian mixture models and other sophisticated background modeling techniques can be considered. 
However, the video streams from UAS have fast motions and no classical background subtraction method would give satisfying results because of the assumption of a static camera. 
Even in a situation where the UAS is hovering over a fixed position, the images are still affected by wind turbulence and vibrations.
Therefore, in order to be able to apply these motion analysis techniques, it is necessary to consider image alignment and video stabilization.
The usual approach is to find strong feature points that can be tracked over a sequence of frames.
Merino \etal \cite{Merino12}, in their fire assistance system, used a motion estimation approach based on feature points matching known as sparse motion field.
From the matched points, they estimate a homography matrix that maps the pixels in an image with the pixels in the previous frame.
This allows mapping every image to a common coordinated frame for alignement.
SURF \cite{bay2008_surf} and ORB \cite{rublee_orb_2011} are two feature point methods that were used for extracting salient features prior to image alignment. 
To our knowledge, the impacts and the benefits of the image alignment has not been yet addressed in the literature relating to fire and smoke detection but some researchers such as \cite{Merino12} consider it important for their fire assistance system to work properly.

\section{Wildland fire datasets}

A large number of fire detection approaches use a classification method that relies on learning algorithms.
The main challenges of machine learning is to build or to find a large enough dataset with low bias.
Such a dataset should contain positive examples with high feature variance and negative examples consisting of standard and challenging samples. 
Deep learning techniques need even larger datasets for training. 
Data augmentation techniques can help in this regard but it requires a sufficiently large dataset to start.

Well-developed research fields such as face or object recognition have already large datasets that have been built and vetted by the community.
These datasets are considered suitable for the development and benchmarking of the new algorithms in their respective fields.
In the case of fire detection, no such widely employed dataset is available yet. But some effort has been made toward this direction.
Steffens \etal \cite{steffens_unconstrained_2015} captured a set of 24 videos from hand-held cameras and robots mounted cameras.
The ground truth was defined by bounding boxes around the fire.
Foggia \etal \cite{foggia_real-time_2015} compiled a collection of 29 videos of fire and smoke but did not provide ground truth data.
In \cite{chino_bowfire:_2015}, the authors gathered around 180 fire images to test their BowFire algorithm and made the dataset available with manually segmented binary images representing the ground truth for the fire area.
However, the main problem with these datasets is the lack of wildfire samples.
This could be problematic for the development of a fire detection module for wildfire assistance systems.
Aerial fire samples in the form of videos are also necessary for the development of UAS based systems. 
In \cite{TOULOUSE2017dataset}, the authors collected images and videos to build the Corsican fire database. This dataset is specifically built for wildfires and forest fires. It also contains multimodal images (visual and NIR images) of fires. The images have their corresponding binary masks representing the ground truth (segmented fire area). Other information are also available such as smoke presence, location of capture, type of vegetation, dominant color, fire texture level, etc.
The dataset contains some aerial forest fire views, but their number is limited. 
The wildfire UAS research is still lacking a dataset that can help improve the development of the algorithms needed in a wildland fire assistance system.
Table \ref{tab:datasets} contains a brief description of the main fire research datasets.


\begin{table}[h!]
\caption{Fire datasets.}
\begin{center}       
\begin{tabular}{|p{3.2cm}|p{3.8cm}|p{2.3cm}|p{2.3cm}|p{2cm}|} 
\hline
\rule[-1ex]{0pt}{3.5ex}  \textbf{Dataset}  &  \textbf{Description} & \textbf{Wildland fires}  &  \textbf{Aerial footage} &  \textbf{Annotations}    \\
\hline
\rule[-1ex]{0pt}{3.5ex}FURG \cite{steffens_unconstrained_2015} & 14,397 fire frames in 24 videos from static and moving cameras. & No & No & Fire bounding boxes \\
\hline
\rule[-1ex]{0pt}{3.5ex}BowFire \cite{chino_bowfire:_2015}&  186 fire and non-fire images. & No & No & Fire masks \\
\hline
\rule[-1ex]{0pt}{3.5ex}Corsican Fire DB \cite{TOULOUSE2017dataset} & 500 RGB and 100 multimodal images. & All & Few  & Fire masks \\
\hline
\rule[-1ex]{0pt}{3.5ex}VisiFire \cite{foggia_real-time_2015}& 14 fire videos, 15 smoke videos, 2 videos containing fire-like objects. & 17 videos & 7 videos & No \\
\hline

\end{tabular}
\end{center}
\label{tab:datasets}
\end{table} 


\section{Fire geolocation and fire modeling}

When a wildland fire is detected, the vehicle must alert the GCS and send the fire location to deploy the firefighting resources.
With a camera located on the bottom of a UAS looking downward, the position of the fire is easy to compute using the UAS GPS position.
For a camera located on the front side of the UAS, we need to project the camera plane onto a global coordinate system using homography.
This transformation maps pixel coordinates to the ground plane.
This approach performs well when the UAS pose estimation is reliable and when the ground is mostly planar.
Some difficulties arise in the presence of uneven surfaces.
Some authors \cite{Martinez-de-Dios07, Ambrosia11, Pastor11, Merino12, Merino15} have circumvented this limitation by exploiting a previously known Digital Elevation Map (DEM) of the surveyed area.
DEM allows for the estimation of the location from where a ray corresponding to a fire pixel originated and thus improves the fire location estimation.
DEMs can induce some errors. To reduce these errors, a UAS fleet looking at the same hotspot can first detect the fire and then use different views of the UAS to refine the estimations \cite{Martinez-de-Dios07}.


Fire models are used to adjust the measurements, to find fire boundaries and to describe the fire behavior.
Simple models use an elliptic shape fitted to fire, where each ellipse axis increases at some given rate.
Other models can be very complex depending on the considered variables.
Such models try to estimate the rate of spread (ROS) of the fire based on wind speed and direction, terrain slopes, vegetation density, weather and other variables.

Ghamry \etal \cite{Ghamry16b, Ghamry16c, Ghamry16d} applied an elliptical model to estimate the fire perimeter.
Here, the rate at which the axis grows depends on the direction towards which the wind blows and its speed. 
The authors assumed that the fire spread more quickly along that direction.
The estimated perimeter is then used to define a UAS team formation.

More complex dynamic models have been used for simulation purposes.
In \cite{Kumar11, Pham17, Lin15}, the authors used the FARSITE (Fire Area Simulation) tool to test their coordination strategies under various scenarios.
The proposed models are not suitable for real-time fire estimation because their complexity significantly increases the computation time.

Some authors use different approaches to model and characterize the fire.
For example, Martinez-de-Dios \etal \cite{Martinez-de-Dios11} use multiple images to extract geometric features from the fire such as the base perimeter, the height and the inclination.
The extraction is performed using computer vision techniques (e.g. image segmentation). 

Bradley \etal \cite{Bradley11} divide the environment into cells and assigns a fire probability is assigned to each cell using IR images.
This method takes into account the uncertainty in the position of the UAS and therefore applies a Gaussian weighting scheme to the probabilities.
The authors then apply a Sequential Monte Carlo (SMC) method to compose a Georeferenced Uncertainty Mosaic (GUM) which is then used to locate the fire.
Belbachir \etal \cite{Belbachir15} model the fire as a static cone of heat sourcing from the fire center and dissipating with an altitude and a horizontal distance.
Based on this assumption, they construct a grid of fire probabilities with the temperature measures. The fire is detected when the probabilities are above a defined threshold.
Lin \etal \cite{Lin15} also generates an occupancy grid with associated temperatures. They also compute the gradient of the grid and estimate the fire center, ROS, and perimeter.

\section{Coordination strategy}
\label{sec:coord}

Coordination strategy is important when deploying autonomous UAS. component. 
It deals with communication, task allocation and planning procedures.
Based on the communication links established during the mission, three main schemes can be distinguished.
First, for a single vehicle, the only possible connection is with the GCS.
For multiple UAS, the path planning and task allocation need to be optimally resolved.
The usual approach is to centralize the heavy processing in the GCS and only allow the UAS to communicate with it but not between each other.
The last scheme tackles the problem of coordinating multiple entities in a distributed manner.
Each vehicle can connect with the GCS and other UAS, allowing for distributed decision-making at a global scale (GCS) and locally (each UAS).

\subsection{Single UAS}

Single UAS, either large airships or small aerial systems, have been evaluated for the viability of using UAS for wildfire surveillance and monitoring.
The Ikhana UAS \cite{Ambrosia11} was deployed in western US between 2006 and 2010.
It was a single large and high endurance vehicle with powerful sensory systems for autonomous fire detection.
The decision strategy and the path planning were performed by human operators.
Pastor \etal \cite{Pastor11} proposed a semi-autonomous system in which a single UAS would sweep a rectangular area, locate hotspots and then return to a nearby ground station.
A human could control the UAS and order it to stay over the hotspot location to confirm visually if it corresponds to a real fire or not.
Martins \etal \cite{Martins07} used an entirely autonomous navigation system where the UAS only received waypoints from where to start surveillance.
When a hotspot is detected, the UAS approaches the source, hover over the target and confirm the fire.
The experimental tests showed very interesting results for fire detection and monitoring tasks.

However, the use of a single UAS is limited when dealing with large-scale wildfires. 
For this reason, the majority of recent developments deal with team-based systems that help increase the area coverage.

\subsection{Centralized team}

The addition of more UAS to the mission increases the area covered by the systems.
In a centralized team strategy, all UAS are coordinated by a single GCS.
This scheme can lead to a more accurate fire georeferencing and reduce false alarms.
In \cite{Martinez-de-Dios07}, the authors follow this approach. They merge data from multiple UAS to correct and reduce the uncertainty of fire georeferencing.
After a fire is detected by a unit, nearby vehicles are sent to the same region to perform a fire confirmation.
Belbachir \etal \cite{Belbachir15} propose a greedy algorithm for fire detection using a probability grid.
For this purpose, each UAS select in a greedy way, the path that provides more information. The UAS visit cells that have not yet been checked and which are within the direction where the temperature increases.
Kumar \etal \cite{Kumar11} propose a coordination protocol for monitoring and firefighting.
In the first step, the planned path of each UAS is optimized to minimize the distance to a detected fire perimeter.
The second step optimizes the path of UAS carrying fire suppressants by minimizing the distance to the fire center.
The last two methods have not been tested in a practical situation, but offer an insight on how a centralized team could work.
The other advantage of centralized processing is that smaller and more affordable UAS can be used as they do not require high-processing power.

\subsection{Collaborative team}

In a distributed fleet, UAS communicate together and with the GCS, allowing the integration of more complex strategies.
The system is able to perform more tasks in an autonomous manner and even to cover larger areas by using some UAS as communication relays.
However, the added complexity imposes new challenges.
Distributed coordination algorithms need to be developed and implemented.
In the literature, these systems were used for optimal fire perimeter surveillance and task allocation.

In \cite{Alexis09}, the authors describe a UAS rendezvous based consensus algorithm which aims to equally distribute the path length of the UAS around the fire perimeter.
UAS depart in pairs and in opposite directions around the fire perimeter.
They set rendezvous locations where they share knowledge about the traveled paths, the current state of the fire perimeter and other units encountered.
If the update shows that the fire perimeter has evolved, then each UAS will select new rendezvous locations in such a way that the distance traveled by each of the UAS is almost the same.
The authors have shown through simulation that the algorithm converge and the recomputing of rendezvous points allows efficient adaptation of the UAS formation to an evolving fire perimeter.

The optimal distribution of UAS around a fire perimeter has also been studied by in \cite{Casbeer06}.
They demonstrated that in order to reduce the latency in communication with the GCS, the UAS must depart in pairs, travel in opposite directions and be evenly spaced around the perimeter.
To achieve optimal perimeter tracking, they designed a control loop to keep half of the bottom-facing IR camera over hotspot pixels and the other half over non-fire area.
Ghamry \etal \cite{Ghamry16c} also distribute the UAS uniformly around the fire perimeter modeled with an elliptical form.
This allows the UAS to keep their paths at even angles around the estimated fire center.
Sujit \etal \cite{Sujit07} proposed the same approach to perform fire perimeter monitoring, but in \cite{Ghamry16d}, the authors added the ability to restructure the formation if a UAS is damaged or has to leave for refueling.
To achieve this fault-tolerant capacity when a UAS needs to leave the formation, all its communications are stoped.
Other vehicles will notice the missing UAS and will start performing the reformation process.
In this system, prior to the monitoring task, the fleet flies in a leader-follower formation where the leader gets a predetermined path plan and the rest follow it at specific distances and angles.
In the work of Lin \etal \cite{Lin15}, UAS are directed to fly uniformly around an estimated fire center.
In this approach Kalman filter is used to estimate the fire contour and the fire center movements, allowing the UAS to fly and adapt their formation accordingly.

For monitoring, Pham \etal \cite{Pham17} propose a collaborative system in which UAS are sent to monitor a fire and optimally cover the fire area.
This formation is achieved by detecting neighboring UAS and reducing camera view overlaps while considering the location of the fire front.
A simple way to model this behavior is with the application of a force field-based algorithm that simulates the attraction of a UAS by the fire front and its repeal from the other UAS.
The attraction and repulsion forces are adapted by considering the fire front confidence and the estimated field of view of each UAS.
One problem with this approach is that the visibility reduction induced by smoke is not taken into account which can put the vehicle in a dangerous situation.

Another coordination strategy was proposed to perform optimal task allocation within a team of UAS.
The tasks can be surveillance, monitoring or firefighting.
In \cite{Ghamry17}, the authors propose an auction-based firefighting coordination algorithm.
In this algorithm, a fire is first detected and then the UAS must coordinate themselves to act upon each known fire spot.
To achieve this task, each vehicle generates a bid valued by a cost function of its distance to the fire spot.
In this manner, the UAS with the best offer for the task will be assigned to it.
Sujit \etal \cite{Sujit07} also proposed a similar auction-based collaboration algorithm but with the ability to consider a minimal number of UAS to watch each hotspot.

\section{Cooperative autonomous systems for wildfires}

UAS can play an important role in the detection and monitoring of large wildfires. 
Multiple UAS can collaborate in the extraction of important data and improve firefighting strategies.
Moreover, aerial vehicles can cooperate with unmanned ground vehicles (UGV) in operational firefighting scenarios.

One type of cooperation can consist in the use of UGV to carry small short endurance UAS to detected fire areas and be used as refueling stations.
Ghamry \etal \cite{Ghamry16b} propose such a system, where a coordinated leader-follower strategy is used.
UAS are carried by UGV to a desired location and deployed to explore preassigned areas.
If a UAS detects a fire, an alert is sent to the leading UGV and to the rest of the fleet.
The leader computes new optimal trajectories for the UAS in order to monitor the fire perimeter.
In \cite{Phan08}, the authors present another firefighting collaborative UAS-UGV strategy.
A hierarchical UAS-UGV system composed of a large leading airship and cooperative UAS and UGV is proposed.
When a fire is detected, the vehicles are deployed for fire monitoring.
In this scenario, UAS and UGV are supposed to have the capacity to carry water and combat fire.
The UAS are deployed in an optimal flying formation over the fire front area.
UGV are sent to prevent the fire propagation and limit its spread using water and fire retardants.
Auction based algorithms are implemented to allocate the tasks to each vehicle.
Viguria \etal \cite{Viguria10} also propose the use of task allocation by an auction-based algorithm.
In their framework, the vehicles can perform various tasks such as surveillance, monitoring, fire extinguishing, transportation and acts as a communication relay.
A human or the GCS can generate a list of tasks that need to be fulfilled.
Each robot sends a bid for each task and the one with the best offer wins and can proceed to execute the task.
The offers are based on specific cost functions for each task that consider the vehicle distance, fuel level and capabilities. 

\begin{figure} 
\centering
\includegraphics[width=0.9\textwidth]{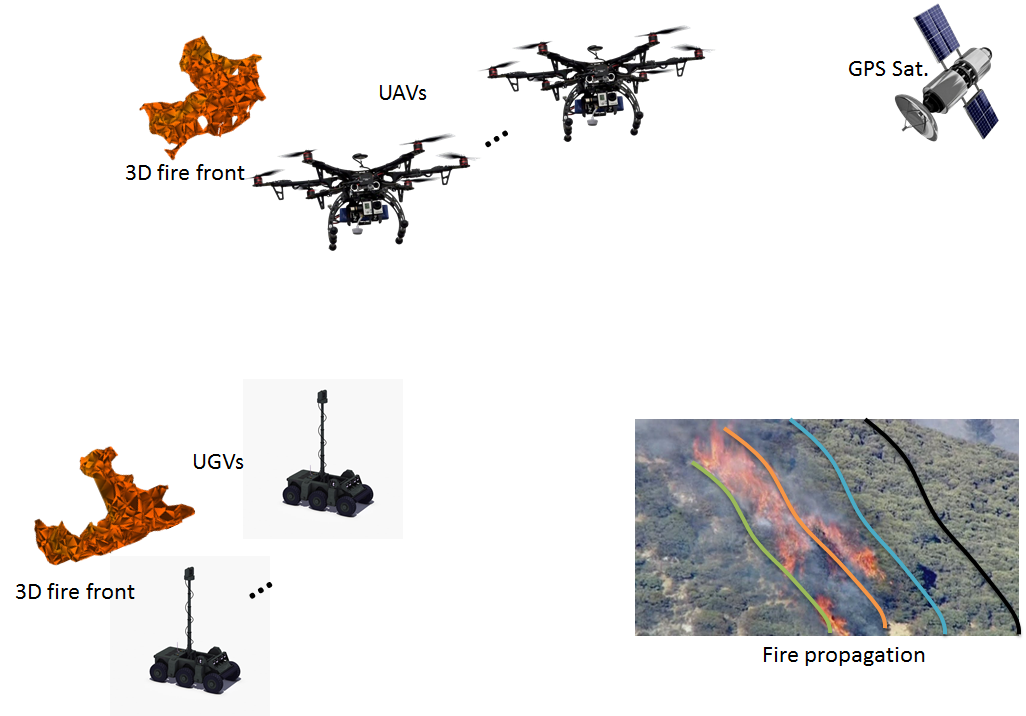}

(a) The acquisition of 3D fire front data; UAS and UGV equipped with multimodal stereo cameras, IMU, and GPS.

\vspace{12pt}

\includegraphics[width=0.9\textwidth]{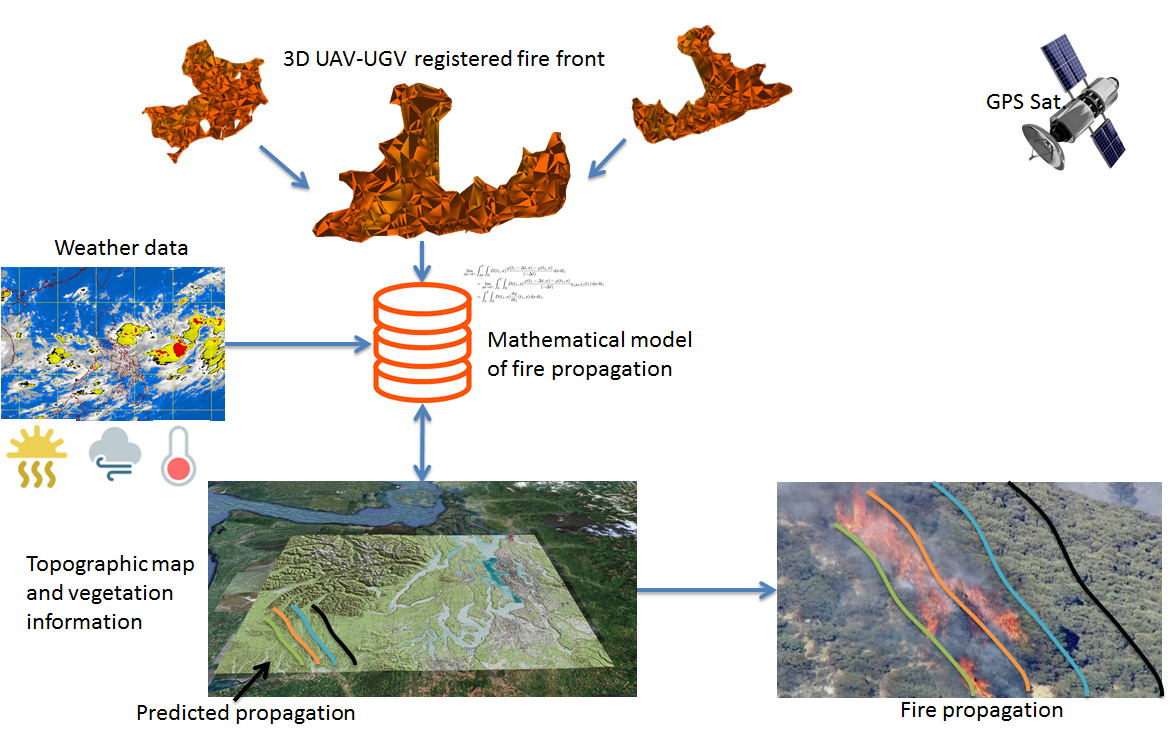}

(b) The modeling and prediction; Registered 3D fire front, weather, topographic and vegetation data are used to predict the fire propagation and map it.

\vspace{12pt}

\caption{UAS-UGV multimodal framework for wildfires assistance~\cite{akhloufiSPIE2018UAVFire}.} 
\label{fig:framework}
\end{figure}

In \cite{akhloufiSPIE2018UAVFire}, the authors propose a multimodal UAS-UGV cooperative framework for large-scale wildfire detection and segmentation, 3D modeling and strategical firefighting. 
The framework is composed of multiple UAS and UGV operating in a team-based cooperative mode. 
Figure \ref{fig:framework} illustrates the proposed framework.
The vehicles are equipped with a multimodal stereo-vision system such as the ones developed for ground-based fire detection and 3D modeling \cite{akhloufi2018MultimodalFire, toulouse2018multimodal, akhloufi2017IPTA, akhloufi2017AITA}.
The stereo system includes multispectral cameras operating in the Visible and NIR spectrum for efficient fire detection and segmentation.
Each stereo system provides an approximate 3D model of the fire. The models captured using multiple views are registered using inertial measurements, geospatial data and the extracted features using computer vision to build the propagating fire front 3D model \cite{akhloufi2018MultimodalFire, toulouse2018multimodal, akhloufi2017IPTA} . 
Based on the 3D model of the fire, the UAS and UGV can be positioned strategically to capture complementary views of the fire front.
This 3D model is tracked over time to compute different three-dimensional fire characteristics such as height, width, inclination, perimeter, area, volume, ROS and their evolution over time.  
The extracted three-dimensional fire characteristics can be fed to a mathematical fire propagation model to predict the fire behavior and spread over time.
The obtained data makes it possible to alert and inform about the risk levels in the surrounding areas.
The predicted fire propagation can be mapped and used in an operational firefighting strategy.
Furthermore, this information can be used for the optimal deployment of UAS and UGV in the field.
This type of framework can be combined with other firefighting resources such as firefighters, aerial firefighting aircraft and future fire extinguisher drones.



\section{Conclusion}
\label{sec:conc}

This paper presents a survey of different approaches for the development of UAS fire assistance systems.
Sensing instruments, fire perception algorithms and different coordination strategies have been described. 

UAS can play an important role in the fight against wildfires in large areas.
With the decrease in their prices and their wider commercial availability, new applications in this field will emerge.
However, some limitations remain such as autonomy, reliability and fault tolerance. Security is also a concern, with the risk associated to having UAS flying over firefighters or close to aircraft carrying water and fire retardants.
Nevertheless, the benefits of using UAS are significant and this could lead to innovations to solve these problems.

In the perception side, most of the developed techniques rely on classical computer vision algorithms. Still, we are witnessing the emergence of some work in the field of deep learning in recent years, especially for fire detection. We presented some of the available datasets containing wildfire images that can be used for developing the computer vision algorithms. However, a small number contain aerial views. In addition, the lack of a large dataset limits the development of advanced deep learning algorithms. Such a dataset can be an important contribution to the field and can serve as a baseline for comparing multiple approaches.  

We also introduced the frameworks proposing cooperative autonomous systems where both aerial and ground vehicles contribute to wildland firefighting. While these frameworks are mostly theoretical and limited to simulation, they provide interesting ideas about a more complete wildfire fighting system. Future research in these areas can provide new approaches for the further development of autonomous operational systems without or with little human intervention.


\section*{Acknowledgements}
This work has been partially supported by the government of Canada under the Canada-Chile Leadership Exchange Scholarship. We acknowledge the support of the Natural Sciences and Engineering Research Council of Canada (NSERC), [funding reference number RGPIN-2018-06233].

\vspace{2cm}

\bibliographystyle{myIEEEtran}    
\bibliography{ms}  

\end{document}